\documentclass[letterpaper, 10 pt, conference]{ieeeconf}  

\IEEEoverridecommandlockouts                              

\usepackage{graphics} 
\usepackage{amsmath} 
\usepackage{amssymb}  

\usepackage[backend=biber,style=ieee,url=false,sortcites=true, mcite=true]{biblatex} 
\usepackage[bookmarks=true]{hyperref}
\AtEveryBibitem{
  \clearfield{isbn}
  \clearfield{issn}
  \clearfield{url}
  \clearfield{urldate}
  \clearfield{doi}
  \clearfield{note}
  \clearfield{editor}
  \clearfield{language}
  \clearfield{langid}
  \clearfield{address}
}
\usepackage{subcaption}
\usepackage{multirow}
\usepackage{pgfplots}
\usepackage{tikz}
\usetikzlibrary{calc}
\usetikzlibrary{shapes.geometric, arrows, positioning}
\usepackage{pdflscape}
\usepackage{censor}
\usepackage[english]{babel}
\usepackage{microtype}
\usepackage{todonotes}
\usepackage{siunitx}
\usepackage{cleveref}

\definecolor{orublue}{HTML}{133455}
\definecolor{orulightblue}{HTML}{3282BE}
\definecolor{orured}{HTML}{BE281B}
\definecolor{orugreen}{HTML}{006F41}
\definecolor{oruyellow}{HTML}{F7A609}
\definecolor{orupurple}{HTML}{732E76}
\definecolor{pointcloud}{HTML}{133455}
\definecolor{alpha}{HTML}{133455}
\definecolor{bpa}{HTML}{732E76}
\definecolor{poisson}{HTML}{BE281B}
\definecolor{vdb}{HTML}{F7A609}
\definecolor{shine}{HTML}{006F41}
\definecolor{3qfp}{HTML}{3282BE}

\title{\LARGE \bf Neural Surface and Reflectance Modelling from 3D Radar Data}

\author{Judith Treffler, Vladimír Kubelka, Henrik Andreasson, Martin Magnusson
\thanks{*This work was partially supported by the Wallenberg AI, Autonomous Systems and Software Program (WASP) funded by the Knut and Alice Wallenberg Foundation, connected to the WASP Research Arena on Public Safety (WARA-PS), and by the EU Horizon Europe Framework Programme (RaCOON project, ID: 101106906). The authors are with ARC and the AASS RNP lab, Örebro University, Sweden {\tt\small \{judith.treffler, vladimir.kubelka, henrik.andreasson, martin.magnusson\}@oru.se}}%
}

\begin{document}

\maketitle
\thispagestyle{empty}
\pagestyle{empty}
\newcommand{\antennaarea}{A_{\mathrm{eff}}}
\newcommand{\transmittedpower}{P_t}
\newcommand{\receivedpower}{P_r}
\newcommand{\signalgain}{G_t}
\newcommand{\sdfloss}{\mathcal{L}_{\mathrm{SDF}}}
\newcommand{\intensityloss}{\mathcal{L}_{\mathrm{I}}}
\newcommand{\meshsampleset}{\mathcal{P}}
\newcommand{\meshsample}{\mathbf{p}_i}
\newcommand{\gtset}{\mathcal{G}}
\newcommand{\gtsample}{\mathbf{g}_i}
\fussy 

\begin{abstract}
Robust scene representation is essential for autonomous systems to safely operate in challenging low-visibility environments. In these conditions, radar has a clear advantage over cameras and lidars due to its resilience to environmental factors such as fog, smoke, or dust. However, radar data is inherently sparse and noisy, making reliable 3D surface reconstruction challenging. To address this, we propose a neural implicit approach for 3D mapping from radar point clouds that jointly models scene geometry and view-dependent radar intensities. Our method leverages a memory-efficient hybrid feature encoding to learn a continuous Signed Distance Field (SDF) for surface reconstruction, while also capturing radar-specific reflective properties. We show that our approach produces smoother, more accurate 3D surface reconstructions compared to existing lidar-based reconstruction methods applied to radar data and can reconstruct view-dependent radar intensities. We also show that, in general, as input point clouds get sparser, neural implicit representations render more faithful surfaces than traditional explicit SDFs and meshing techniques. 
\end{abstract}

\section{Introduction}
\label{sec:introduction}
Accurate and reliable representations of the environment are essential for autonomous systems, particularly in low-visibility conditions. However, classical explicit mapping approaches often struggle with sparse or noisy data. In the last few years, neural implicit scene representations, such as Neural Radiance Fields (NeRF)~\cite{mildenhall_nerf_2021}, have become a widely adopted framework for 3D mapping \cite{isaacson_loner_2023, zhong_shinemapping_2023, sun_3qfp_2024}. Unlike explicit representations, which employ fixed spatial data structures, neural methods implicitly represent the surface structure with a neural network and trainable features, enabling maps with improved surface reconstruction at arbitrary resolutions. Still, the majority of these methods focus on camera or lidar data and might fail in low-visibility environments. Radar, with its ability to penetrate clouds of small particles such as fog, smoke, or dust with minimal attenuation, is a promising alternative. Beyond this robustness, the view-dependent intensities and reflections of radar data can provide valuable information for localisation and scene understanding. However, radar also introduces additional challenges, as the noisy data -- particularly multi-path reflections -- makes accurate modelling more difficult.

Despite its potential, radar-based neural implicit scene representations have remained unexplored until recently. Existing methods focus on 2D novel view synthesis (NVS) and provide only limited 3D reconstruction \cite{borts_radar_2024, huang_dart_2024, zhang_rf4d_2026}, perform NVS of 3D radar point clouds via projections to 2D range maps \cite{sheng_georf_2025}, or can only model radar jointly with camera and lidar \cite{rafidashti_neuradar_2025}. To the best of our knowledge, no prior work has reconstructed 3D surfaces from radar point clouds while modelling view-dependent radar intensities.

\begin{figure}[t]
    \centering
    \includegraphics[width=0.95\linewidth]{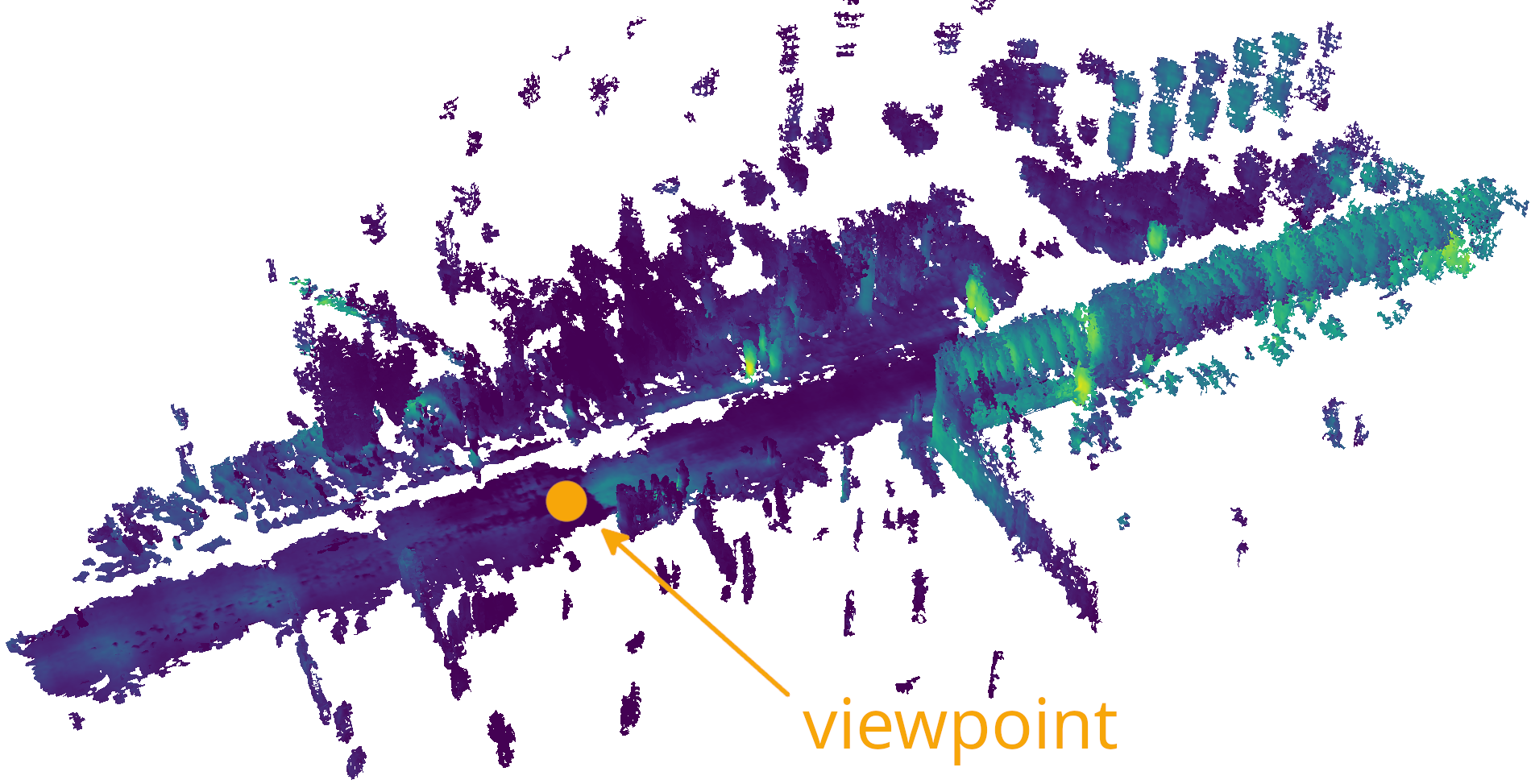}
    \caption{Surface reconstruction of the \texttt{Viking Hill} dataset, coloured according to the predicted view-dependent radar intensities for the viewpoint marked in yellow.}
    \label{fig:surface-reconstruction-result}
\end{figure}

In this paper, we propose a neural implicit approach for 3D mapping from radar point clouds that offers both accurate surface reconstruction and modelling of view-dependent reflectance. We base our method on the memory-efficient 3QFP representation, designed for dense lidar point clouds \cite{sun_3qfp_2024}, to learn a Signed Distance Field (SDF) for surface reconstruction directly from 3D radar data. Inspired by the architecture of NeuS2~\cite{wang_neus2_2023}, we introduce an intensity network that models view-dependent radar intensities, enabling our method to learn both geometric and reflective aspects from radar data. We further analyse the impact of NeuS2 architecture components on the intensity reconstruction quality when applied to radar.
 
Our main contributions are as follows:
\begin{itemize}
    \item A memory-efficient learnable radar reflectivity model that accounts for surface geometry and view-dependent backscatter, improving reconstruction quality and robustness in low-visibility and feature-poor environments.
    \item An evaluation of 3D reconstruction methods on two radar sensors with different noise characteristics.
    \item A discussion on metrics for assessing radar-based environment modelling.
\end{itemize}

As our experiments demonstrate, our approach produces more accurate surface reconstructions with smoother locally planar regions from radar data, compared to both classical explicit methods and recent neural representations, while also predicting realistic view-dependent radar intensities across varying viewing angles.
\section{Related Work}
\label{sec:related_work}
3D mapping has traditionally relied on explicit scene representations, such as occupancy grids~\cite{hornung_octomap_2013, kramer_radarinertial_2021}, surfels~\cite{behley_efficient_2018,wang_realtime_2019}, meshes~\cite{edelsbrunner_shape_1983, bernardini_ballpivoting_1999, kazhdan_poisson_2006, vizzo_poisson_2021, han_online_2023} or Truncated Signed Distance Field (TSDF) values~\cite{oleynikova_voxblox_2017, vizzo_vdbfusion_2022}. These discretised spatial representations often face challenges with scalability, memory inefficiency at high resolutions, and reconstruction of fine details or unobserved areas. In the last few years, neural implicit methods have gained popularity. In contrast to explicit approaches, they use multi-layer perceptrons (MLPs) to learn a continuous scene representation, enabling reconstruction at different resolutions, improved completion, and more robustness to sparse input data. The seminal work introducing Neural Radiance Fields (NeRF)~\cite{mildenhall_nerf_2021} learns a volumetric scene representation from 2D images. While intended for novel view synthesis, scene geometry can be extracted via volume rendering; however, since surface geometry is inferred from the volume density field, high-quality surface extraction remains challenging and often yields noisier reconstructions.

To learn surface representations more directly, several approaches learn implicit occupancy fields~\cite{mescheder_occupancy_2019, oechsle_unisurf_2021, yan_efficient_2023} or signed distance functions~\cite{yariv_volume_2021}. In particular, NeuS \cite{wang_neus_2021} and its extension NeuS2 \cite{wang_neus2_2023} combine volume rendering with an SDF and use a separate network to model view-dependent components. Beyond camera-based methods, SHINE-Mapping~\cite{zhong_shinemapping_2023} and 3QFP~\cite{sun_3qfp_2024} learn an SDF from lidar data. SHINE-Mapping uses an octree-based data structure to store learnable features. Building on this, 3QFP reduces memory usage while maintaining reconstruction quality by replacing octrees with more efficient tri-quadtrees and using Fourier feature positional encoding. Neural implicit scene representations have further been applied to other sensors, including ultrasound~\cite{wysocki_ultranerf_2023}, sonar~\cite{qadri_neural_2023, xie_bathymetric_2024}, or synthetic aperture radar (SAR)~\cite{lei_sarnerf_2024, sugavanam_neural_2024}. Most neural implicit reconstruction methods, however, focus on RGB, RGB-D, lidar, or combinations thereof and might fail in low-visibility environments. While several advances have been made to overcome these challenges~\cite{mildenhall_nerf_2022, chen_dehazenerf_2024}, radar offers an inherently more robust solution. Its longer wavelength enables radar signals to penetrate clouds of small particles (dust, smoke, fog, etc.) with little attenuation, making it well-suited for low-visibility conditions without specialised correction techniques. For autonomous systems, two types of mmWave radar sensors are commonly used: spinning radars that capture 2D polar images or system-on-a-chip (SoC) radars that generate 3D radar data cubes with an additional velocity dimension, often referred to as 3+1D or 4D radar.

Despite these advantages, the sparse and noisy nature of radar data poses challenges for neural implicit representations, which have only recently been explored. Initial work with 2D radars focuses on adapting NeRF to radar images for NVS. DART~\cite{huang_dart_2024} performs NVS on range-Doppler images, whereas Radar~Fields~\cite{borts_radar_2024} and RF4D~\cite{zhang_rf4d_2026} perform NVS and 3D mapping from 2D radar scans and are thus limited in elevation and detail in their scene reconstruction. For 3D radar, GeoRF~\cite{sheng_georf_2025} reconstructs and refines point clouds but relies on learning 2D range maps from down-projected 3D data, which can lead to information loss. NeuRadar~\cite{rafidashti_neuradar_2025} instead jointly models camera, lidar, and 3D radar data for NVS, but cannot operate using radar alone. While these methods explore related directions, they address different tasks or operate on different sensor measurements or modalities and are therefore unsuitable as direct baselines for our work. Consequently, to the best of our knowledge, no existing work reconstructs surfaces from only radar point clouds. Furthermore, while Radar~Fields and RF4D model 2D radar reflectance and combine it with occupancy for NVS, modelling radar intensities has not yet been extended to 3D radar data, leaving this source of scene information unexplored. Since radar reflections encode view-dependent and material-specific properties, explicitly modelling intensities provides complementary information to the reconstructed geometry, motivating our approach to 3D radar-based surface and reflectance modelling.

\section{Method}
\label{sec:method}
We model the scene geometry by learning a continuous SDF and view-dependent radar intensities with an SDF and an intensity network. The weights of both networks, together with feature vectors stored in a memory-efficient tri-quadtree grid, are jointly optimised using radar range detections and intensities from known sensor poses. The learned SDF allows for mesh reconstruction using marching cubes~\cite{lorensen_marching_}, while the intensity network can be queried at arbitrary 3D points and viewing directions to reconstruct view-dependent radar intensities. We will refer to our architecture as 3QFPI (3QFP with intensities).

\subsection{Network}
\begin{figure}
    \centering
    \resizebox{\linewidth}{!}{
        \begin{tikzpicture}[node distance=2cm]
    \tikzstyle{block} = [rectangle, 
            draw, 
            thick,
            fill=orulightblue!, 
            text centered, 
            minimum height=1.5cm, 
            minimum width=2cm, 
            rounded corners=3pt, 
            align=center]
    \tikzstyle{layerblock} = [rectangle, 
                draw, 
                thick,
                fill=orulightblue!60, 
                text centered, 
                minimum height=1.25cm, 
                minimum width=0.7cm, 
                rounded corners=1pt, 
                align=center]
    \tikzstyle{feature_block}=[draw, thick, fill=orulightblue!60, rounded corners=1pt] 
    \tikzstyle{arrow} = [thick,->,>=latex,rounded corners=1pt]
    \tikzstyle{line} = [thick,rounded corners=1pt]
    \tikzstyle{bluearrow} = [thick,->,>=latex,orulightblue,dashed,rounded corners=1pt]
    \tikzstyle{yellowdots} = [thick,oruyellow,dotted,rounded corners=1pt]

    \node (origin) at (0,0){};
    \node (feature) [feature_block, minimum width=0.5cm, minimum height=2cm, right=0.5cm of origin, yshift=0cm] {};
    \node (feature-text) [left=0.25cm of feature.north, yshift=-0.3cm]{\parbox{2cm}{\centering tri-quadtree\\ \vspace{-1mm}feature}};
    \node (fourierfeature) [feature_block, minimum width=0.5cm, minimum height=0.8cm, below=9mm of feature] {};
    \node (fourierfeature-text) [below=0mm of fourierfeature]{\parbox{4cm}{\centering Fourier feature\\\vspace{-1mm}positional encoding}};
    \node (sdf-network) [block, right=1cm of feature, yshift=-1.02cm] {};
    \node (int-label) [below=-0cm of sdf-network] {SDF network};
    \node (sdf-l1) [layerblock, right=1.2cm of feature, yshift=-1.02cm] {\scriptsize 32};
    \node (sdf-l2) [layerblock, right=2.1cm of feature, yshift=-1.02cm] {\scriptsize 32};
    \node (intensity-network) [block, right= 2cm of sdf-network] {};
    \node (int-label) [below= 0cm of intensity-network] {\parbox{2cm}{\centering intensity\\\vspace{-0.1cm}network}};
    \node (int-l1) [layerblock, right=2.2cm of sdf-network] {\scriptsize 64};
    \node (int-l2) [layerblock, right=3.1cm of sdf-network] {\scriptsize 64};
    \node (loss)[right=1.5cm of intensity-network] {\parbox{1.5cm}{\centering \textcolor{orulightblue}{
    $\mathcal{L}_{\mathrm{SDF}}$ \\
    \vspace{-0.15cm}
    + \\
    \vspace{-0.15cm}
    $\mathcal{L}_{\mathrm{Intensity}}$
    }}};
    \coordinate (view-dir-temp) at ($(intensity-network.south west) -(1cm,0.75cm)$);
    \node (view-dir) [feature_block, minimum width=1.5cm, minimum height=0.5cm] at ($(intensity-network.south west) -(1.5cm,2.3cm)$) {};
    \node (view-dir-text) [left=0.25cm of view-dir.west, yshift=0.05cm] {\parbox{1.5cm}{\centering viewing\\ \vspace{-1mm}direction}};
    \node (sh) [below=0.1cm of view-dir] {\parbox{1.5cm}{\centering Spherical\\ \vspace{-1mm}Harmonics}};

    \coordinate (sdf_point1) at ($(sdf-network.north east) + (-0.5cm,0)$);
    \coordinate (sdf_point2) at ($(intensity-network.north west) + (0.5cm,0.5cm)$);
    \coordinate (feature-arrow-point) at ($(sdf-network.west) + (-0.5cm,0)$);
    \coordinate (fourier-feature-point) at ($(fourierfeature.east) + (0.5cm,0)$);
    \coordinate (3dpoint-arrow-point1) at ($(feature.west) + (-0.5cm,0)$);
    \coordinate (3dpoint-arrow-point2) at ($(fourierfeature.west) + (-0.5cm,0)$);
    \coordinate (3dpoint-arrow-midpoint) at ($(3dpoint-arrow-point1)!0.5!(3dpoint-arrow-point2)$);
    \coordinate (pointcloud-coord) at ($(3dpoint-arrow-midpoint) +(-0.5cm,0)$);
    \coordinate (backprop-arrow-loss) at ($(loss.north west) - (0,0.18cm)$); 
    \coordinate (backprop-arrow-int) at ($(backprop-arrow-loss) - (1.5cm,0)$);
    \coordinate (backprop-arrow-midpoint) at ($(backprop-arrow-int)!0.5!(backprop-arrow-loss)$);
    \coordinate (temp) at ($(feature.north east) - (0,-0.2cm)$);
    \coordinate (backprop-arrow-feature) at ($(temp |- backprop-arrow-midpoint)+(0,1.35cm)$);
    
    \draw[yellowdots] (feature-text.north west) -- ($(feature-text.north west -| sdf-network.east) + (0.2cm,0)$) -- ($(sdf-network.east |- fourierfeature-text.south west) + (0.2cm,0)$) -- node[pos=0.925, above, yshift=-1mm] {3QFP} ($(feature-text.north west |- fourierfeature-text.south west)$) -- cycle;

    \coordinate (3qfpi_corner1) at ($(feature-text.north west) + (-0.2cm,0.2cm)$);
    \coordinate (3qfpi_corner2) at (3qfpi_corner1 |- sh.south);
    \draw[yellowdots, orugreen] (loss.east |- 3qfpi_corner1) -- (3qfpi_corner1) -- (3qfpi_corner2) -- node[pos=0.035, above, yshift=-1mm] {3QFPI} (3qfpi_corner2 -| loss.east) -- cycle;
    
    \draw [line] (feature) -- ++(0.75cm,0) -- (feature-arrow-point);
    \draw [arrow] (feature-arrow-point) -- (sdf-network.west);
    \draw [line,rounded corners=0pt] (fourierfeature.east) -- ++(0.5cm,0) -- (feature-arrow-point);
    \draw [line] (fourier-feature-point) -- ++(4.22cm,0);
    \draw [arrow] (sdf-network) --(intensity-network) node[midway, above, yshift=-0.9cm]{\parbox{1.5cm}{\centering \textcolor{black!55}{geometry\\ \vspace{-1mm}feature \\ \vspace{2mm} SDF\\ \vspace{-1mm}normals}}};
    \draw [arrow] (intensity-network) -- (loss) node[midway, below, yshift=0mm] {intensity};
    \draw [line] (sdf_point1) -- ++(0, 0.5cm) -- (sdf_point2) node[midway, above, yshift=-0.4mm]{SDF};
    \draw [arrow] (sdf_point2) -- ++(0, -0.5cm);
    \draw [arrow] (sdf_point2) --(loss.north |- sdf_point2) -- ($(loss.north) + (0,0.1cm)$);
    \node (pointcloud)[left=-0.15cm of pointcloud-coord] {\parbox{1cm}{\centering 3D\\ \vspace{-1mm}point}};
    \draw [line] (pointcloud-coord) -- (3dpoint-arrow-midpoint);
    \draw [arrow] (3dpoint-arrow-midpoint) -- (3dpoint-arrow-point1) -- (feature.west);
    \draw [arrow] (3dpoint-arrow-midpoint) -- (3dpoint-arrow-point2) -- (fourierfeature.west);
    \draw [line] (view-dir) -- ($(view-dir.east) + (0.45cm,0)$) -- ($(intensity-network.west) + (-0.3cm,0)$);
    \draw [arrow] (view-dir.west) +(-0.4cm,0) -- (view-dir.west);

    \draw [bluearrow] (backprop-arrow-loss) -- (backprop-arrow-int);
    \draw [bluearrow] (backprop-arrow-midpoint) -- ++(0,1.35cm) -- (backprop-arrow-feature);
    \draw [bluearrow] ($(sdf-network.north)+(0,0.9cm)$) -- (sdf-network.north);
\end{tikzpicture}
    }
    \caption{\textbf{Network Architecture}: Given a 3D point $\textbf{x}$, we concatenate its tri-quadtree feature and Fourier feature positional encoding and pass them to the SDF network. The SDF network predicts an SDF value and, optionally, a learned geometry feature and/or approximated SDF normals. These outputs, along with the spherical harmonics-encoded viewing direction, and the tri-quadtree and Fourier feature encodings of $\textbf{x}$, are concatenated and fed into the intensity network to predict the intensity value for $\textbf{x}$.}
    \label{fig:architecture}
\end{figure}

Inspired by NeuS2~\cite{wang_neus2_2023}, we separate geometry and reflectance modelling into an SDF and an intensity network. The intensity network is conditioned on the scene geometry learned by the SDF network. \Cref{fig:architecture} shows an overview of the architecture.

\subsubsection{SDF Network}	
We use 3QFP~\cite{sun_3qfp_2024} as our SDF network due to its memory-efficient yet powerful hybrid feature encoding. 3QFP encodes each 3D point with tri-quadtrees by projecting a 3D point $\textbf{x}$ onto three axis-aligned orthogonal planes and constructing a quadtree on each plane. Nodes at the deepest $H$ levels of a quadtree contain learnable feature vectors for interpolation. These vertex features are stored in hash tables to enable fast queries. The learnable tri-quadtree feature encoding is combined with Fourier feature positional encoding, creating a hybrid representation that produces smoother results and improves completion, with only a minimal increase in computational cost~\cite{sun_3qfp_2024}. For further details on 3QFP's architecture, we refer to the original paper~\cite{sun_3qfp_2024}.

To integrate 3QFP into the NeuS2 architecture, we modify the network to optionally output a geometry feature vector $\textbf{g}\in \mathbb{R}^{15}$ and SDF normals $\textbf{n}$, alongside the SDF value $d$. The geometry feature and the normals, calculated as the gradients of the SDF $\textbf{n}=\nabla_\textbf{x}d$, provide additional geometric information to the intensity network. Both outputs are optional, as our ablation study (\cref{subsec:ablation}) showed minimal effect on the intensity reconstruction quality.

\subsubsection{Intensity Network}
The intensity network predicts the intensity of a radar return at a 3D point in space, given a viewing direction. According to the radar equation,
\begin{equation}
    \receivedpower = \frac{\transmittedpower \signalgain \antennaarea \sigma}{(4\pi)^2 r^4}
\end{equation}
this received intensity depends on the object properties and distance to the sensor. The transmitted power~$\transmittedpower$, signal gain~$\signalgain$, and effective area of the receiver~$\antennaarea$ are constant for all detected objects. If the manufacturer provides the information, these constants are known; here, the network learns them implicitly. The radar cross section $\sigma$ depends on the incident angle (viewing direction), angle of reflection, material and object size, and can be inferred from the scene geometry or learned implicitly.

Our intensity network takes the 3D point $\textbf{x}$, the viewing direction $\textbf{v}$, the SDF value $d$, and optionally the geometry feature $\textbf{g}$ and SDF normals $\textbf{n}$ as input. Consistent with the SDF network, we encode $\textbf{x}$ with tri-quadtree and Fourier feature positional encoding, while $\textbf{v}$ is encoded with Spherical Harmonics. From these inputs, the network predicts the radar intensity at $\textbf{x}$ when observed from $\textbf{v}$. 

\subsection{Training}
\subsubsection{Sampling}
We adopt the sampling process from 3QFP, sampling both near the input radar points and in free space. For every input point $\textbf{x}$, we randomly sample $N_s$ points along the ray near $\textbf{x}$, and $N_f$ points along the ray in free space between the sensor and $\textbf{x}$. Both SDF and intensity labels are assigned to the same sampled points. The SDF labels are set as the signed distance between the sampled point and $\textbf{x}$. Samples between the sensor and $\textbf{x}$ are assigned negative labels; samples beyond $\textbf{x}$ have positive labels. For the intensity labels, we assign the same intensity as the ground-truth point to samples near $\textbf{x}$, assuming local intensities are approximately constant due to the low spatial resolution of radar. We normalise the ground-truth intensities to the range [0,1] to compensate for sensor-specific thresholding. Since we do not expect a radar return in free space, we set the label of free-space samples to 0. 

\subsubsection{Loss Function}
The loss function consists of an SDF component and an intensity component. For the SDF loss, we follow SHINE-Mapping and 3QFP~\cite{zhong_shinemapping_2023, sun_3qfp_2024}: We first map the SDF value to [0,1] with a sigmoid function and then use a Binary Cross-Entropy loss. This gives sample points near the surface a higher impact and allows faster convergence. As with 3QFP, we do not include additional SDF regularisation, Eikonal or loss weight terms, since they did not noticeably affect the reconstruction. For the intensity loss, we use an L1 loss, which was experimentally found to achieve more accurate results than an L2 loss, presumably because it is less sensitive to outliers in sparse, noisy radar data.
\section{Experiments}
\label{sec:experiments}
\subsection{Experiment Setup}
\subsubsection{Baselines}
We compare our surface reconstruction to meshes created with classical mesh generation methods: $\alpha$-shapes~\cite{edelsbrunner_shape_1983}, Ball-Pivoting Algorithm (BPA)~\cite{bernardini_ballpivoting_1999}, and Poisson Surface Reconstruction (Poisson)~\cite{kazhdan_poisson_2006}. Additionally, we evaluate against VDBFusion~\cite{vizzo_vdbfusion_2022}, a TSDF-based explicit scene reconstruction method, and SHINE-Mapping~\cite{zhong_shinemapping_2023}, a lidar-based neural implicit mapping approach that learns an SDF. 

As an intensity reconstruction baseline, we implement a simple k-nearest-neighbour-inspired approach (kNN) that considers both the spatial proximity and similarity in viewing direction. For each query point, we first identify a set of nearest-neighbour candidates based on spatial distance. From this set, we select the k points with the smallest viewing-angle difference and estimate the intensity as a weighted average of their intensities, where the weights depend on both the spatial distance and viewing-angle difference.

\begin{table*}[htb]
   \vspace{0.1cm} 
   \centering
   \caption{Quantitative evaluation of the surface reconstruction quality on the \texttt{Viking Hill} and \texttt{SNAIL-Radar} datasets using lidar point clouds as ground truth. We compare accuracy and completion errors (cm), and accuracy ratio, completion ratio and F-score (\%) with a threshold of \SI{0.2}{\m}. Distances above \SI{0.4}{\m} are omitted from the accuracy calculation, and the accuracy outlier ratio is additionally reported. To assess local planarity of surfaces, we evaluate the shape, mean, and variance of the Gamma distribution fitted to the histogram of angles between adjacent triangles (see \cref{fig:angle-histogram}). Overall, 3QFPI achieves the best accuracy and preserves planar surface patches, while Poisson and SHINE-Mapping achieve the best completion. Best results per metric are highlighted in \textbf{bold}; second-best results \underline{underlined}.}
   \setlength{\tabcolsep}{4.5pt}
   \renewcommand{\arraystretch}{1.1}
   \begin{tabular}{c c| c c c c c c c c c}
       \hline
       \multirow[c]{2}{*}{dataset} & \multirow[c]{2}{*}{method} &  Acc. &  Acc. & Acc. Out. & Comp. & Comp. & \multirow[c]{2}{*}{F-Score (\%) $\uparrow$} & Gamma & Gamma & Gamma\\
       & & Error (cm) $\downarrow$ &  Ratio (\%) $\uparrow$ & Ratio (\%) $\downarrow$ & Error (cm) $\downarrow$ & Ratio (\%) $\uparrow$ & & Shape & Mean $\downarrow$ & Variance $\downarrow$ \\ 
       \hline
       \multirow[c]{6}{*}{\shortstack{\texttt{Viking}\\\texttt{Hill}}} & $\alpha$-shapes & 15.66 & 65.41 & 37.48 & \underline{30.95} & 66.22 & 65.81 & 1.35 & 0.86 & 0.55\\
       & BPA & 15.95& 64.99 & 34.21 & 34.93  & 61.34 & 63.11 & 1.75 & 0.65 & 0.24\\
       & Poisson & 15.90 & 65.07 & 48.99 & \textbf{25.18} & \textbf{74.64} & \underline{69.53} & 1.09 & \underline{0.49} & \underline{0.22}\\
       & VDBFusion & 14.93 & 69.18 & \textbf{23.85} & 40.76 & 55.73 & 61.73 & 1.20 & 0.54 & 0.24\\
       & SHINE & \underline{14.80} & \underline{69.32} & 32.10 & 31.70 & \underline{66.42} & 67.84 & 1.27 & 0.53 & 0.22\\
       & 3QFPI & \textbf{13.36} & \textbf{73.62} & \underline{26.16} & 34.53 & 66.23 & \textbf{69.73} & 0.72 & \textbf{0.31} & \textbf{0.13}\\  
       \hline
       \multirow[c]{6}{*}{\shortstack{\texttt{SNAIL-}\\\texttt{Radar}\\Basketball\\Court}} & $\alpha$-shapes & 18.06 & 59.24 & 61.44 & 30.83 & 50.21 & 54.35 & 2.24 & 1.22 & 0.67\\
       & BPA & 16.79 & 63.57 & 45.12 & 38.93 & 44.27 & 52.20 & 1.81 & 0.69 & 0.26\\
       & Poisson & 17.51 & 61.13 & 59.28 & \underline{18.19} & \textbf{76.30} & \underline{67.87} & 1.14 & \underline{0.52} & \underline{0.24}\\
       & VDBFusion & \underline{16.76} & \underline{63.80} & \underline{44.70} & 33.51 & 56.83 & 60.11 & 1.52 & 0.67 & 0.30\\
       & SHINE & 17.12 & 62.72 & 50.67 & \textbf{17.14} & \underline{75.99} & \textbf{68.72}& 1.42 & 0.59 & 0.25\\
       & 3QFPI & \textbf{16.75} & \textbf{64.59} & \textbf{44.43} & 46.73 & 41.56 & 50.58 & 0.91 & \textbf{0.44} & \textbf{0.21}\\ 
       \hline
   \end{tabular}
   \label{tab:mesh_quality_comparison}
\end{table*}

\begin{figure*}[htb]
    \centering
    \input{figures/plots/meshes}
    \caption{Surface reconstruction quality of different methods on the \texttt{Viking Hill} dataset (\subref{fig:forest-rgb}--\subref{fig:mesh-forest-3qfp}), and a corner of the \texttt{SNAIL-Radar} basketball court dataset showing a basket and a building in the background (\subref{fig:snail-rgb}--\subref{fig:mesh-snail-3qfp}). For reference, we include an image of the scene (from a different angle) and the lidar-based reconstruction created with SHINE-Mapping; meshes are coloured by surface normals. The comparison indicates that 3QFPI produces more accurate and smoother locally planar surfaces from noisy data. In particular, the reconstruction of the building in (\subref{fig:mesh-snail-3qfp}) is closest to the lidar reference (\subref{fig:mesh-snail-lidar}).
    \vspace{-0.1cm}
    }
    \label{fig:mesh_quality_comparison}
\end{figure*}

\subsubsection{Evaluation Metrics}
We evaluate the surface reconstruction quality following the setup and thresholds from \cite{zhong_shinemapping_2023, sun_3qfp_2024} by uniformly sampling $\text{10}^7$ points on each mesh and calculating the \textit{accuracy} and \textit{completion errors}. The \textit{accuracy error} is the average minimum distance from the mesh samples to ground-truth lidar points, discarding distances above \SI{0.4}{\m}. We additionally report the percentage of discarded distances as the \textit{accuracy outlier ratio}. The \textit{completion error} is computed as the average minimum distance from ground-truth points to the mesh samples, with distances truncated at \SI{2.0}{\m}. Lidar point clouds are used as ground truth, since they are denser and less noisy than radar data. We also report the \textit{accuracy} and \textit{completion ratio}, and \textit{F-score} as the percentage of errors below a threshold of \SI{0.2}{\m}. To account for the different fields of view and ranges of radar and lidar, we manually restrict the evaluation to the bounding box where both sensors overlap and have high point density.

The distance thresholds applied in the evaluation can substantially affect the results. In particular, discarding larger distances in the accuracy calculation can lead to an overestimation of the accuracy of meshes with strong interpolation in unobserved areas. We report the aforementioned metrics for comparison with lidar-based reconstructions, but note that they do not fully capture reconstruction quality for sparser radar data.
 
To assess reconstruction of locally planar surfaces from noisy data, we fit a Gamma distribution to the histogram of angles between adjacent triangles (see \cref{fig:angle-histogram}), reporting shape, mean, and variance in \cref{tab:mesh_quality_comparison}. Lower mean and variance indicate smoother surface patches, while lower shape values correspond to distributions concentrated near zero with a long tail of larger angles at edges or sharp features. We note that these metrics are supplementary and should be considered alongside previously described metrics and qualitative evaluation, as their interpretation depends on the scene. In environments like those in our datasets, where the majority of surfaces are planar, lower mean and variance should correspond to higher-quality reconstructions.

To evaluate the predicted intensities, we hold out every tenth point cloud during training and compare the reconstructed intensities to the ground truth using the Mean Absolute Error (MAE) and Median Absolute Error (MedAE) in \cref{tab:intensity-reconstruction-errors}, \cref{tab:ablation}, and \cref{fig:view-dependent-intensity-reconstruction}.

\subsubsection{Datasets}
We evaluate our method on three different outdoor radar datasets: 1) the \texttt{Viking Hill} dataset\footnote{Dataset available at \url{https://github.com/RNP-lab/viking_hill_radar_lidar_camera_dataset}} using the first segment of the short grass run (Hugin A3, 48Tx/48Rx antennas); 2) the publicly available \texttt{SNAIL-Radar} dataset~\cite{huai_snail_2025}, using the denser AI-enhanced radar point clouds from the Basketball Court (20231007/4) and Starlake Tower (20231208/1) sequences (Oculii Eagle, 6Tx/8Rx antennas); and 3) a dataset specifically recorded to assess view-dependent intensity recovery, consisting of three sequences: Wall with Window 1, Wall with Window 2, and Wall without Window. The dataset was collected by moving the sensor in a semicircle around a corner reflector mounted on a wall to capture measurements at varying incidence angles and approximately constant range (Hugin A4, 48Tx/48Rx antennas).

We remove points within a \SI{2.5}{\m} radius (\SI{1}{\m} for \texttt{SNAIL-} \texttt{Radar}) around the sensor to suppress near-field clutter and densify point clouds by accumulating five consecutive frames.

\subsubsection{Implementation}
We implement the classical surface reconstruction methods using Open3D~\cite{zhou_open3d_2018}, setting the parameters to achieve reasonable performance on our datasets: $\alpha$-shapes with $\alpha$\,=\,0.2; BPA with a pivoting radius of \SI{0.1}{\m}; and Poisson with an octree depth of 12. For VDBFusion, we set the voxel size to \SI{0.2}{\m} and enable isolated vertex filtering. SHINE-Mapping and 3QFPI use a leaf node resolution of \SI{0.2}{\m}. For mesh extraction and rendering of VDBFusion, SHINE-Mapping, and 3QFPI, we use a voxel size of \SI{0.1}{\m}. In 3QFPI, we use the same hyperparameter settings as \cite{sun_3qfp_2024}, but increase the number of free-space and near-surface samples to 6 to improve completion, train for 4000 iterations with a learning rate of $\text{10}^{-3}$, and freeze the SDF network after 1000 iterations. We normalise intensities using the rounded dataset-wide minimum and maximum for both the training and test data.

\subsection{Surface Reconstruction Experiment Results}
\subsubsection{Surface Reconstruction with Dense Input Data}
\begin{figure}[!tb]
    \centering
    \begin{subfigure}[t]{0.5\linewidth}
        \centering
        \raisebox{0pt}{\input{figures/plots/angle-histogram-forest}}
        \caption{\texttt{Viking Hill}}
        \label{fig:angle-histogram-forest}
    \end{subfigure}\hspace{-0.2cm}
    \begin{subfigure}[t]{0.5\linewidth}
        \centering
        \raisebox{0pt}{\input{figures/plots/angle-histogram-snail-basket}}
        \caption{\texttt{SNAIL-Radar}}
        \label{fig:angle-histogram-snail}
    \end{subfigure}
    \caption{Outlines of the histograms of angles between adjacent mesh triangles for surface reconstructions from the \texttt{Viking Hill} and \texttt{SNAIL-Radar} basketball court datasets. For both datasets, 3QFPI produces the largest proportion of small angles, indicating better preservation of locally planar regions, such as at the ground or walls.}
    \label{fig:angle-histogram}
\end{figure}

We first evaluate the surface reconstruction using all frames of each dataset. \Cref{tab:mesh_quality_comparison} shows that 3QFPI achieves the lowest accuracy error and highest accuracy ratio on both datasets, indicating close geometric alignment with the lidar ground truth, albeit at the cost of reconstruction completeness on the \texttt{SNAIL-Radar} dataset. 
While VDBFusion yields a slightly lower accuracy outlier ratio on the \texttt{Viking Hill} dataset, likely because its sparser reconstruction results in fewer radar artefacts, 3QFPI preserves more radar details but also artefacts, which increase the outlier count. Poisson surface reconstruction achieves the highest completion ratio and competitive F-score; however, the reconstructions are overly smooth (see \cref{fig:mesh-forest-poisson,fig:mesh-snail-poisson}), as the method fills unobserved areas to generate watertight surfaces. This interpolation can introduce geometrically incorrect regions and lead to a high outlier rate. SHINE-Mapping achieves high completion and accuracy but preserves more radar details and noise, also resulting in an increased outlier ratio.

The qualitative evaluation with \texttt{SNAIL-Radar} data in \cref{fig:mesh_quality_comparison} indicates that a high F-score does not necessarily correspond to a geometrically accurate reconstruction. Since mesh samples more than \SI{0.4}{\m} from the nearest ground-truth point are excluded from the accuracy calculation -- as in SHINE-Mapping and 3QFP~\cite{zhong_shinemapping_2023, sun_3qfp_2024} -- reconstructions with a high outlier ratio can still achieve high F-scores. This effect seems more pronounced in radar-based reconstruction, due to the sparse and noisy data. As a result, both Poisson and SHINE-Mapping achieve high F-scores, while their over-smoothed or noisy reconstructions obscure details, such as the building or the basketball goal (see \cref{fig:mesh-snail-poisson,fig:mesh-snail-shine}). In contrast, 3QFPI tends to extract planar surfaces, such as the ground or building facades, more reliably from noisy data, producing meshes that are more consistent with the lidar reference in planar areas. This observation is supported by the distribution of angles between adjacent triangles in the mesh (see \cref{fig:angle-histogram}), where 3QFPI yields the most angles close to zero for both datasets. As \cref{tab:mesh_quality_comparison} shows, the fitted Gamma distribution further quantifies this behaviour: 3QFPI produces the lowest mean and variance, indicating smoother and more homogeneous local surface regions, and a lower shape parameter, reflecting more nearly coplanar triangles.

\subsubsection{Surface Reconstruction with Sparse Input Data}
We evaluate robustness to sparse input using every $\text{n}^{\text{th}}$ point cloud from the \texttt{Viking Hill} dataset as input. As shown in \cref{fig:sparse-input}, Poisson consistently achieves the highest completion ratios across all sparsity levels, due to its watertight surfaces that interpolate unobserved areas. However, as this interpolation can over-smooth surfaces, high completion does not necessarily reflect reconstruction quality. The completion of other classical meshing approaches declines more rapidly. In contrast, the neural implicit methods SHINE-Mapping and 3QFPI remain more robust to sparse input, likely because the continuous scene representation can infer unobserved regions.

\begin{figure}[!tb]
\begin{minipage}[b]{0.235\textwidth}
    \centering
    \raisebox{0pt}{\hspace{-1.9mm}\begin{tikzpicture}
    \tikzset{every node/.append style={font=\tiny}}
        \begin{axis}[
            line width=0.8pt,
            scale only axis=false,
            width=1.225\linewidth,
            height=1.5\linewidth,
            xlabel={Every $\text{n}^{\text{th}}$ Frame},
            ylabel={Completion Ratio (\%)},
            xlabel style={yshift=4mm},
            ylabel style={yshift=-7.5mm},
            xmin=0.7, xmax=10.3,
            ymin=0, ymax=76,
            xtick={1,2,3,4,5,6,7,8,9,10},
            ytick={10,20,30,40,50,60,70},
            xticklabel style={yshift=0.5mm},
            yticklabel style={xshift=0.5mm},
            xtick pos=bottom,
            ytick pos=left,
            legend pos=south west,
            legend style={
                row sep=-2pt,
                draw=black!80,
                line width=0.3pt,
                xshift=-0.25mm,
                yshift=-0.5mm,
            },
            legend image post style={
                line width=0.8pt
            },
            ymajorgrids=true,
            grid style=dashed,
            axis line style={line width=0.3pt, color=black!80},
        ]
        \addplot[
            color=alpha,
            mark=square,
            ]
            coordinates {
            (1,66.2152676297783)(2,62.7156418042681)(3,58.7818556171389)(4,54.5409580177533)(5,51.5885181510669)(6,48.0152762796964)(7,46.1540235880647)(8,41.8111880138346)(9,38.4601048737046)(10,36.6504371485152)
            };
        \addplot[
            color=bpa,
            mark=pentagon,
            ]
            coordinates {
            (1,61.3386062884905)(2,53.5918889979594)(3,48.752123489377)(4,42.2974968971695)(5,40.0608718331142)(6,34.4601599186384)(7,32.3258057202171)(8,28.8231524233532)(9,26.1729223486363)(10,25.8991681662042)
            };
        \addplot[
            color=poisson,
            mark=star,
            ]
            coordinates {
            (1,74.6448094593408)(2,72.0595795877659)(3,69.605589944077)(4,66.393547685295)(5,64.377026341622)(6,62.1770795517246)(7,61.1649867040273)(8,58.1468074593749)(9,57.1949626022099)(10,53.7526228255613)
            };
        \addplot[
            color=vdb,
            mark=triangle,
            ]
            coordinates {
            (1,55.7332837055201)(2,51.0706239605354)(3,47.2644371721059)(4,43.4600590029266)(5,38.5175612997419)(6,35.5488438598032)(7,29.9984010757353)(8,27.3157207020064)(9,26.441148447012)(10,21.5760281934908)
            };
        \addplot[
            color=shine,
            mark=diamond,
            ]
            coordinates {
            (1,66.4164856955255)(2,66.338319268479)(3,65.2577243113813)(4,63.6295580809239)(5,61.9985688317237)(6,59.3298672499682)(7,56.1804451562424)(8,54.6426994035488)(9,50.9383588484076)(10,49.9308032759599)
            };
        \addplot[
            color=3qfp,
            mark=o,
            ]
            coordinates {
            (1,66.2282687188816)(2,64.6182044080507)(3,62.7792055967592)(4,61.3675389541201)(5,58.812549720647)(6,55.3592664869406)(7,52.2736571985012)(8,51.4319546324899)(9,47.3420374226914)(10,47.6333037576031)
            };
            \legend{$\alpha$-shapes, BPA, Poisson, VDBFusion, SHINE-Mapping, 3QFPI}
        \end{axis}
\end{tikzpicture}}
    \captionof{figure}{Comparison of completion ratios using every $\text{n}^{\mathrm{th}}$ point cloud from the \texttt{Viking Hill} dataset as input. Except for Poisson, the completion of classical meshing methods declines quickly with sparse input, whereas neural implicit methods remain more robust.}
    \label{fig:sparse-input}
\end{minipage}
\hfill
\begin{minipage}[b]{0.235\textwidth}
    \centering
    \raisebox{0pt}{\hspace{-1.9mm}\begin{tikzpicture}
    \tikzset{every node/.append style={font=\tiny}}
    \begin{axis}[
        line width=0.8pt,
        scale only axis=false,
        width=1.22\linewidth,
        height=1.5\linewidth,
        xlabel={\# Input Frames},
        ylabel={Map Size (MB)},
        xlabel style={yshift=3.26mm},
        ylabel style={yshift=-6.5mm},
        xmin=8, xmax=120,
        ymin=0, ymax=300,
        xtick={11,23,39,58,117},
        ytick={0,50,100,150,200,250},
        xticklabel style={yshift=0.5mm},
        yticklabel style={xshift=0.5mm},
        xtick pos=bottom,
        ytick pos=left,
        legend pos=north west,
        legend style={
            row sep=-2pt,
            draw=black!80,
            line width=0.3pt,
            xshift=-0.25mm,
            yshift=0.5mm,
        },
        legend image post style={
            line width=0.8pt
        },
        ymajorgrids=true,
        grid style=dashed,
        axis line style={line width=0.3pt, color=black!80},
    ]
        \addplot[
            color=pointcloud,
            mark=square,
            ]
            coordinates {
            (11,22.6)(23,43.5)(39,71.5)(58,108.6)(117,290.1)
            };
        \addplot[
            color=vdb,
            mark=triangle,
            ]
            coordinates {
            (11,10.7)(23,16.8)(39,22.2)(58,26.6)(117,35.9)
            };
        \addplot[
            color=shine,
            mark=diamond,
            ]
            coordinates {
            (11,36.1)(23,52.4)(39,62.4)(58,73.8)(117,91.2)
            };
        \addplot[
            color=3qfp,
            mark=o,
            ]
            coordinates {
            (11,12.1)(23,14.8)(39,16.5)(58,18.2)(117,20.7)
            };
            \legend{point clouds, VDBFusion, SHINE-Mapping, 3QFPI}
    \end{axis}
\end{tikzpicture}}
    \captionof{figure}{Comparison of memory usage for the \texttt{Viking Hill} dataset. 3QFPI has the lowest memory usage, requiring only about $20-30\%$ of the map size of SHINE-Mapping; however, SHINE-Mapping captures more fine details.}
    \label{fig:memory-usage}
\end{minipage}
\end{figure}

\subsubsection{Memory Usage}
We compare the memory usage of different map representations to that of point clouds. Following \cite{sun_3qfp_2024}, we use the stored model parameters (feature embeddings and MLP weights) as the map size for SHINE-Mapping and 3QFPI, and the VDB structure storing TSDF values for VDBFusion. The memory required to store extracted meshes is not included in this comparison. As \cref{fig:memory-usage} shows, 3QFPI has the lowest memory usage, increasing only slightly with the number of input frames due to its efficient tri-quadtree representation. In contrast, dense point cloud maps require the most memory. While 3QFPI achieves a completion ratio comparable to SHINE-Mapping on the \texttt{Viking Hill} dataset, it requires considerably less memory, likely because SHINE-Mapping stores features in a hierarchical octree structure and reconstructs more noise and fine details. Compared to VDBFusion, 3QFPI reconstructs more geometric details (see \cref{fig:mesh_quality_comparison}) while using slightly less memory.

\subsection{Intensity Reconstruction Experiment Results}
\subsubsection{Intensity Reconstruction Quality}
\begin{table}[!tb]
    \centering
    \caption{Evaluation of intensity reconstruction quality. We compare the MAE and MedAE for 3QFPI and a kNN-inspired baseline across different scenes with and without view-dependent intensities. Results for 3QFPI are averaged over 10 repetitions. The errors are expressed on the same scale as the radar intensity measurements. The low errors relative to the sensor's dynamic range indicate that our method accurately reconstructs view-dependent intensities.}
    \setlength{\tabcolsep}{3pt}
    \renewcommand{\arraystretch}{1.1}
    \begin{tabular}{c| c|c | c|c |c}
        \hline
        \multirow{2}{*}{dataset} & \multicolumn{2}{c|}{3QFPI} & \multicolumn{2}{c|}{kNN} & intensity\\
        & MAE & MedAE & MAE & MedAE & range\\
        \hline
        \texttt{Viking Hill} & 3.24 $\pm$\,0.04 & 2.38 $\pm$\,0.03 & 3.64 & 2.60 & 64\,--\,118\\
        Wall w/ Window 1 & 2.64 $\pm$\,0.04 & 2.00 $\pm$\,0.02 & 3.96 & 2.95 & 80\,--\,138\\
        Wall w/ Window 2 & 2.48 $\pm$\,0.03 & 1.82 $\pm$\,0.02 & 3.12 & 2.95 & 80\,--\,128\\
        Wall w/o Window & 2.79 $\pm$\,0.04 & 2.09 $\pm$\,0.03 & 3.32 & 2.47 & 80\,--\,133\\
        Basketball Court & 4.92 $\pm$\,0.08 & 2.72 $\pm$\,0.06 & 4.44 & 3.50 & 0\,--\,35\\
        Starlake Tower & 4.50 $\pm$\,0.10 & 2.74 $\pm$\,0.04 & 4.25 & 3.28 & 0\,--\,36\\
        \hline
    \end{tabular}
    \label{tab:intensity-reconstruction-errors}
\end{table}

\begin{figure*}[!tb]
    \centering
    \begin{subfigure}[t]{0.24\textwidth}
        \centering
        \includegraphics[width=\textwidth]{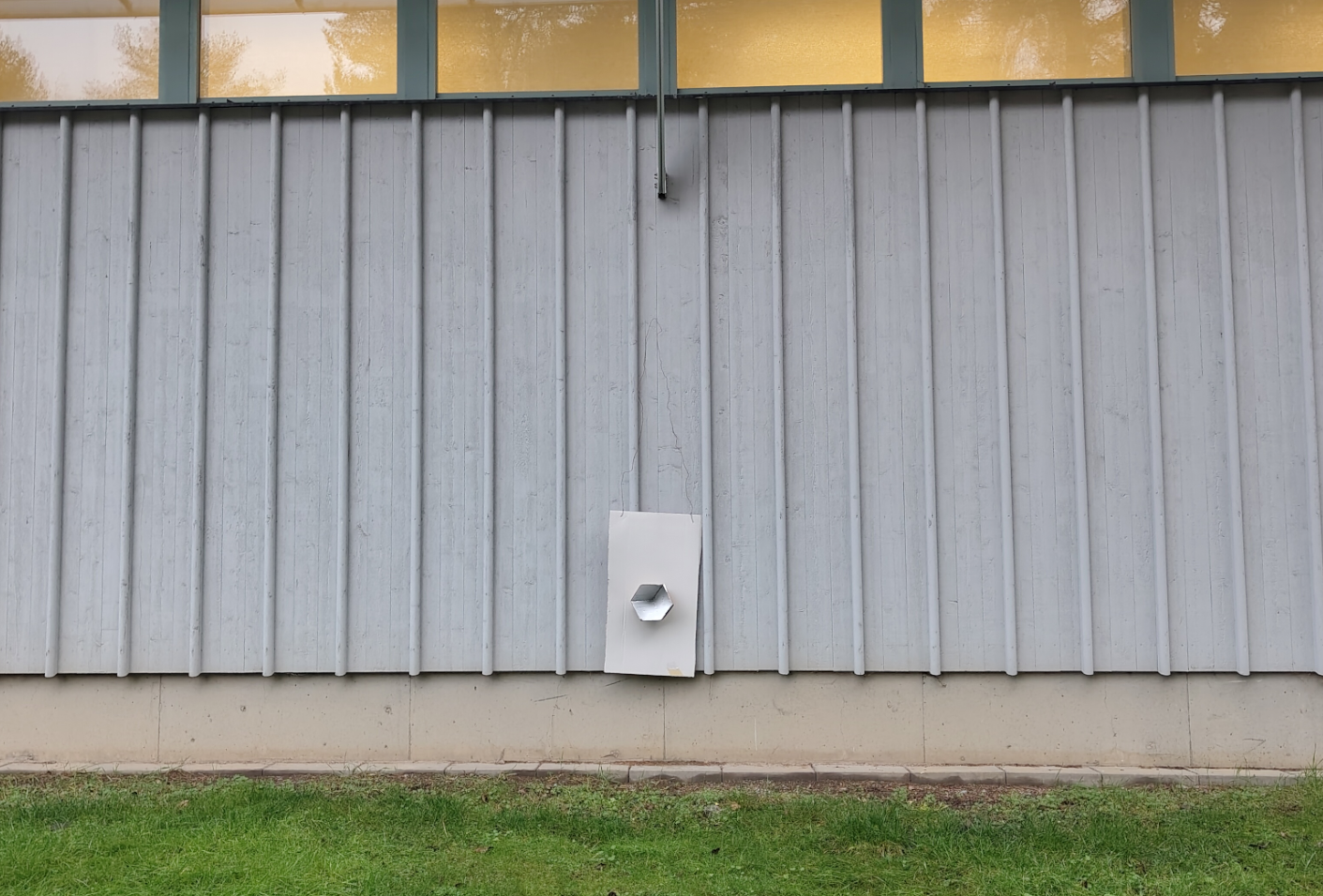}
        \caption{corner reflector}
        \label{subfig:view-dependent-reflector}
    \end{subfigure}
    \begin{subfigure}[t]{0.24\textwidth}
        \centering
        \includegraphics[width=\textwidth]{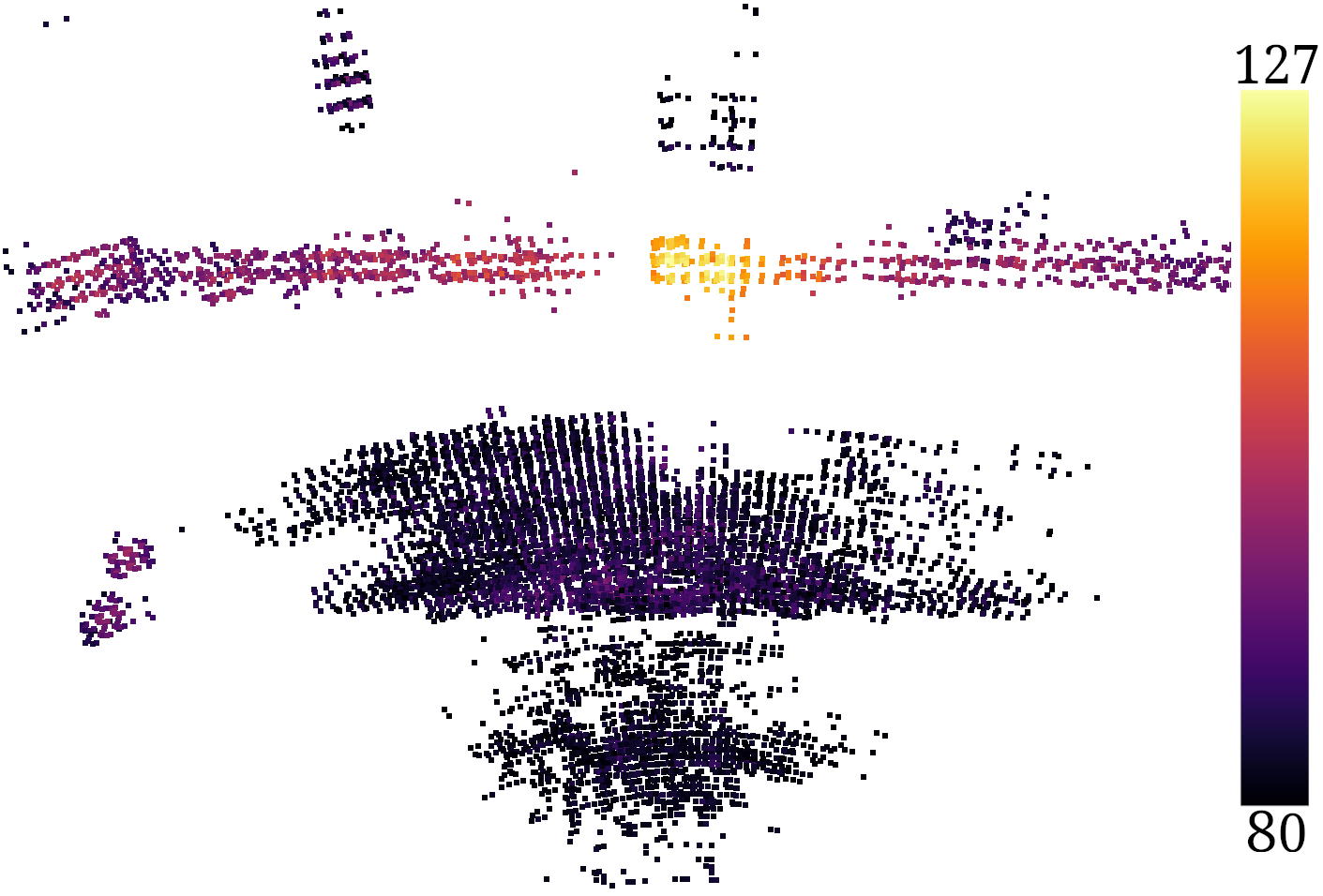}
        \caption{gt intensities}
        \label{subfig:view-dependent-gt}
    \end{subfigure}
    \begin{subfigure}[t]{0.24\textwidth}
        \centering
        \includegraphics[width=\textwidth]{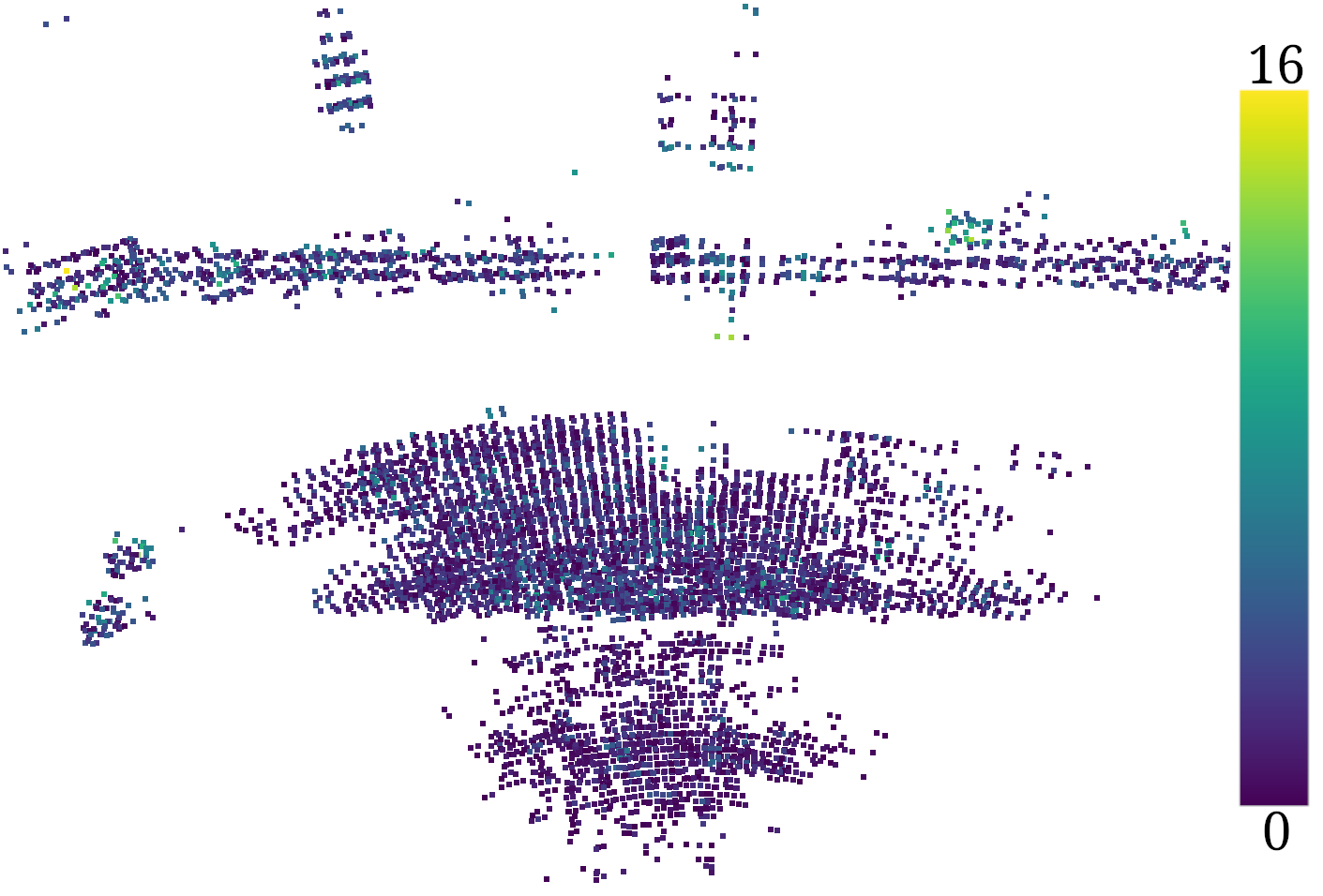}
        \caption{\centering reconstruction errors\\full network}
        \label{subfig:view-dependent-errors}
    \end{subfigure}
    \begin{subfigure}[t]{0.24\textwidth}
        \centering
        \includegraphics[width=\textwidth]{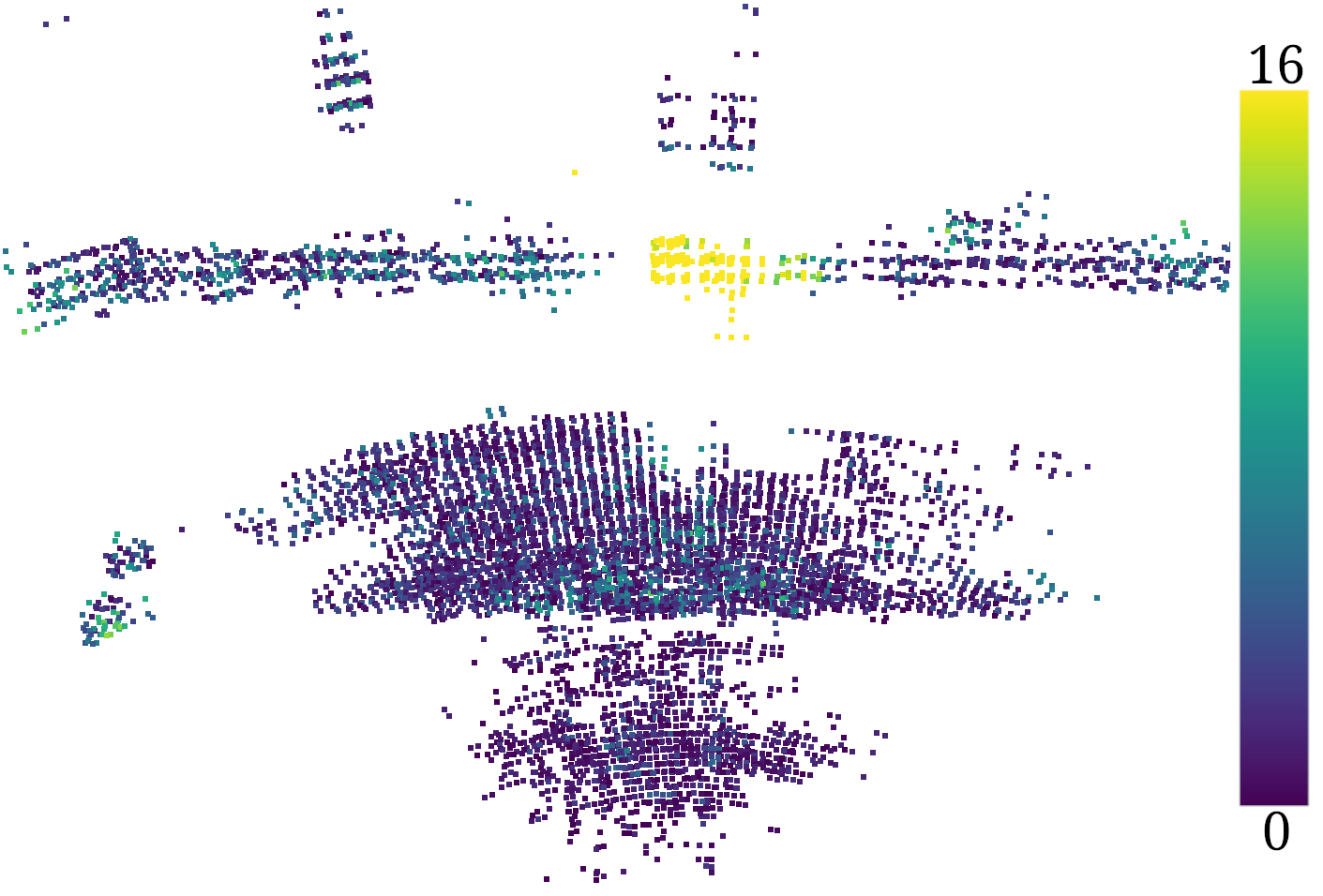}
        \caption{\centering reconstruction errors\\w/o viewing direction}
        \label{subfig:view-independent-errors}
    \end{subfigure}
    \caption{Reconstruction of view-dependent intensities. A corner reflector mounted on a wall (\subref{subfig:view-dependent-reflector}) is observed from multiple viewing directions in the "Wall w/o Window" dataset. The measured intensities of points on the wall decrease with lower incidence angles (\subref{subfig:view-dependent-gt}). Our model reconstructs these view-dependent intensities with low per-point reconstruction errors (\subref{subfig:view-dependent-errors}), whereas the ablation without viewing-direction input produces larger errors, particularly around the corner reflector (\subref{subfig:view-independent-errors}).}
    \label{fig:view-dependent-intensity-reconstruction}
\end{figure*}

To evaluate the intensity reconstruction quality, we compare the predicted intensities from 3QFPI and a kNN-inspired baseline for test frames against the corresponding ground-truth point clouds. As \cref{tab:intensity-reconstruction-errors} shows, our method predicts intensities with lower errors than the kNN-inspired baseline on all Hugin datasets. The MAE of 3QFPI's predictions remains small relative to the dynamic range of the sensors, corresponding to 5-6\% of the intensity range for the Hugin radar in \texttt{Viking Hill} and the view-dependent data, and 13-14\% for the Oculii radar in \texttt{SNAIL}. \Cref{fig:view-dependent-intensity-reconstruction} further illustrates that our method accurately predicts the decrease in intensity with lower incident angles. While the baseline achieves lower MAE on the \texttt{SNAIL} datasets, the lower MedAE for 3QFPI suggests this is caused by larger outliers. More generally, the higher reconstruction errors on the \texttt{SNAIL-Radar} datasets suggest that reconstruction accuracy is sensitive to the quality of the sensor data.

\subsubsection{Ablation Study}
\label{subsec:ablation}
\begin{table*}[!tb]
   \centering
   \caption{Ablation study on the inputs to the intensity network. We report MAE and MedAE (mean $\pm$ standard deviation over 10 repetitions) for different input configurations. Removing the viewing direction or SDF input noticeably increases the intensity reconstruction errors, while other ablations have only a minor effect.
   The errors are expressed on the same scale as the radar's intensity measurements (see \cref{tab:intensity-reconstruction-errors} for intensity ranges).
   }
   \setlength{\tabcolsep}{2pt}
   \renewcommand{\arraystretch}{1.1}
   \begin{tabular}{c| c|c | c|c | c|c | c|c | c|c}
        \hline
        \multirow{2}{*}{dataset} & \multicolumn{2}{c|}{full} &  \multicolumn{2}{c|}{w/o SDF} & \multicolumn{2}{c|}{w/o SDF normals} &  \multicolumn{2}{c|}{w/o geometry feature} & \multicolumn{2}{c}{w/o viewing direction}\\
        & MAE & MedAE & MAE & MedAE & MAE & MedAE & MAE & MedAE & MAE & MedAE \\
        \hline
        \texttt{Viking Hill} & 3.24 {$\pm\,0.04$} & 2.38 $\pm\,0.03$ & 3.35 $\pm\,0.04$ & 2.44 $\pm\,0.03$ & 3.23 $\pm\,0.07$ & 2.37 $\pm\,0.04$ & 3.25 $\pm\,0.06$ & 2.39 $\pm\,0.04$ & 3.80 $\pm\,0.05$ & 2.75 $\pm\,0.04$ \\
        Wall w/ Window 1 & 2.64 $\pm\,0.04$ & 2.00 $\pm\,0.02$ & 2.69 $\pm\,0.03$ & 2.04 $\pm\,0.02$ & 2.67 $\pm\,0.03$ & 2.01 $\pm\,0.02$ & 2.63 $\pm\,0.03$ & 2.00 $\pm\,0.02$ & 4.05 $\pm\,0.05$ & 3.04 $\pm\,0.04$ \\
        Wall w/ Window 2 & 2.48 $\pm\,0.03$ & 1.82 $\pm\,0.02$ & 2.54 $\pm\,0.02$ & 1.84 $\pm\,0.02$ & 2.48 $\pm\,0.04$ & 1.82 $\pm\,0.03$ & 2.49 $\pm\,0.03$ & 1.82 $\pm\,0.02$ & 3.16 $\pm\,0.02$ & 2.15 $\pm\,0.02$ \\
        Wall w/o Window & 2.79 $\pm\,0.04$ & 2.09 $\pm\,0.03$ & 2.85 $\pm\,0.04$ & 2.13 $\pm\,0.04$ & 2.83 $\pm\,0.04$ & 2.11 $\pm\,0.03$ & 2.82 $\pm\,0.06$ & 2.11 $\pm\,0.05$ & 3.42 $\pm\,0.04$ & 2.55 $\pm\,0.04$ \\
        Basketball Court & 4.92 $\pm\,0.08$ & 2.72 $\pm\,0.06$ & 5.08 $\pm\,0.05$ & 2.78 $\pm\,0.04$ & 4.91 $\pm\,0.10$ & 2.72 $\pm\,0.09$ & 4.97 $\pm\,0.06$ & 2.75 $\pm\,0.06$ & 5.12 $\pm\,0.03$ & 2.86 $\pm\,0.03$ \\
        Starlake Tower & 4.50 $\pm\,0.10$ & 2.74 $\pm\,0.04$ & 4.84 $\pm\,0.04$ & 2.86 $\pm\,0.04$ & 4.50 $\pm\,0.13$ & 2.73 $\pm\,0.07$ & 4.60 $\pm\,0.08$ & 2.76 $\pm\,0.06$ & 4.75 $\pm\,0.03$ & 2.90 $\pm\,0.04$ \\
        \hline
   \end{tabular}
   \label{tab:ablation}
\end{table*}

We analyse the impact of different inputs to the intensity network on reconstruction quality by removing the SDF network, SDF normals, geometry feature, and viewing direction. \Cref{tab:ablation} shows that training only the intensity network leads to higher reconstruction errors, emphasising the importance of learning scene geometry separately. However, omitting the SDF normals or geometry feature has only a minor effect, suggesting that 3D radar data, unlike RGB images, already provides sufficient geometric context for view-dependent intensity estimation. Removing viewing direction causes the most noticeable increase in error across all datasets. As illustrated in \cref{fig:view-dependent-intensity-reconstruction}, this increase is particularly pronounced around specular objects such as the corner reflector. This highlights that modelling radar intensities without accounting for viewing direction is insufficient to capture their strongly view-dependent characteristics.

\section{Conclusion} 
\label{sec:conclusion}
In this paper, we present 3QFPI, a neural implicit approach for 3D scene reconstruction from radar point clouds that jointly models geometry and view-dependent radar intensities. Experiments with two radar sensors indicate that neural implicit representations in general are particularly well-suited for 3D mapping from sparse and noisy radar data, compared to explicit SDF or meshing techniques. Despite the low spatial resolution typical of 3D radars, our method produces dense, smooth, and geometrically consistent reconstructions. In addition, it successfully captures view-dependent radar responses, enabling the model to explain not only where surfaces are, but also how they reflect energy. This joint modelling of geometry and intensity provides a richer and more physically meaningful scene representation than purely geometric reconstructions. However, performance notably depends on the quality of the input data. We further observe that common evaluation metrics may not fully capture reconstruction quality with sparse radar data. To provide a more faithful assessment, we therefore complement conventional metrics (accuracy, completion, F-score) with additional quantitative analyses (local surface smoothness, accuracy outlier ratio) and qualitative inspection, highlighting the limitations of current evaluation practices. Future work will focus on modelling radar-specific characteristics, such as multi-path reflections and the effects of a wide beam, to enable more physically accurate scene reconstruction.



\printbibliography

\end{document}